\documentclass[11pt]{article}

\usepackage[final]{acl}
\usepackage{algorithm}
\usepackage{algpseudocode}
\usepackage{listings}
\usepackage{amssymb}
\lstnewenvironment{promptbox}{\lstset{basicstyle=\ttfamily\scriptsize,
    breaklines=true, breakindent=0pt, columns=fullflexible, keepspaces=true,
    frame=single, framesep=4pt, xleftmargin=4pt, xrightmargin=4pt,
    escapeinside={(*}{*)}}}{}

\usepackage{times}
\usepackage{latexsym}
\usepackage{amsmath}
\usepackage{todonotes}
\usepackage{booktabs}
\usepackage{amssymb}

\usepackage[T1]{fontenc}
\usepackage[utf8]{inputenc}

\usepackage{microtype}

\usepackage{inconsolata}

\usepackage{graphicx}

\title{Guiding AI Agents to Generate Conjectures \\  by Encoding How Theorems Build on One Another}
\author{Hita Kambhamettu \\
  University of Pennsylvania \\
  \texttt{hitakam@seas.upenn.edu} \\\And
  Daniel Krashen \\
  University of Pennsylvania \\
  \texttt{dkrashen@math.upenn.edu} \\\And
  Lyle Ungar \\
  University of Pennsylvania \\
  \texttt{ungar@cis.upenn.edu} \\}
\begin{document}
\maketitle

\begin{abstract}

Recently, LLMs have been making headlines, and headway, in mathematical research. Much of this work focuses on proving existing open conjectures. Deciding what to prove is harder and less studied, because of the lack of a clear criterion for what makes a conjecture worth pursuing. We present a method for conjecture generation that builds on the observation that mathematical theorems build on one another: new theorems can modify the setting or hypotheses of earlier theorems, strengthen their conclusions, or extend their scope. We encode these relationships in a new kind of graph, the Theorem Evolution Graph (TEG), which we instantiate as an online search method for LLM agents. An agent builds a TEG in two steps: (1) identifying relevant theorems from papers found by following citations backward and forward from a seed paper; and (2) comparing those theorems to record how one builds on another. LLM judges and an expert mathematician preferred TEG-generated conjectures over those from an unconstrained baseline. LLM judges favored TEG conjectures for significance, ambition, and insight, and favored baseline conjectures for plausibility. Ablations suggest a benefit from constructing and using a TEG even when the available papers are fixed.

\end{abstract}

\section{Introduction}

Mathematical research has recently become an lively frontier for LLMs \citep{openai2026math,openai2026navierstokes,anthropic2026riemann,anthropic2026fermat}. LLMs are being used for autoformalization \citep{wu2022autoformalization}, for generating natural-language proofs of conjectures \citep{feng2026autonomous}, and for formal theorem proving \citep{hubert2026alphaproof}. This increased interest is in part because mathematics has verification built into it: a proposed result can be supported by a proof which can be formalized, and a formal proof can be checked mechanically. This gives models a signal for evaluating and improving their outputs.

As such, much of the work in AI for math focuses on proving open problems. A comparatively less studied question is: what do we want to prove in the first place? Recent work has begun to generate conjectures as training data for theorem provers \citep{dong2025stp,onda2025leanconjecturer}, but choosing conjectures worth research pursuit is a harder task. What makes a problem ``good'' is harder to concretize. Mathematical correctness alone does not tell us whether a conjecture is novel, significant, or likely to deepen our understanding of a field.

Our intuition for how to generate conjectures with ``good'' properties comes from the structural interconnectedness of mathematics: new theorems often build on earlier ones by modifying hypotheses, strengthening conclusions, or extending results to new settings. Encoding these relationships could suggest fruitful new conjectures.

Our contribution is the \emph{Theorem Evolution Graph} (TEG), a new kind of graph that encodes how mathematical theorems build on one another. Nodes represent theorems, and edges describe what changes from one theorem to another. LLM agents construct and use the TEG during conjecture generation, drawing on a seed paper and related work to propose new conjectures. An agent constructs a TEG through two operations: (1) identify, finding relevant theorems in papers reached by following citations backward and forward from the seed paper; and (2) compare, recording how those theorems build on one another.

We investigate the utility of TEGs in three controlled experiments: we compare TEG agents against an unconstrained baseline, isolate which papers an agent reads versus whether it builds a TEG from them, and ablate the graph's nodes and edges. First, LLM and expert judges prefer conjectures generated by TEG agents over those generated by an unconstrained baseline. Second, ablations suggest a benefit from constructing and using a TEG even when the available papers are held fixed. Third, we change what the graph contains. Conjectures built from the full TEG are judged more novel and insightful than those built from the graph's nodes alone.

\begin{figure}[t]
\centering
\includegraphics[width=\linewidth]{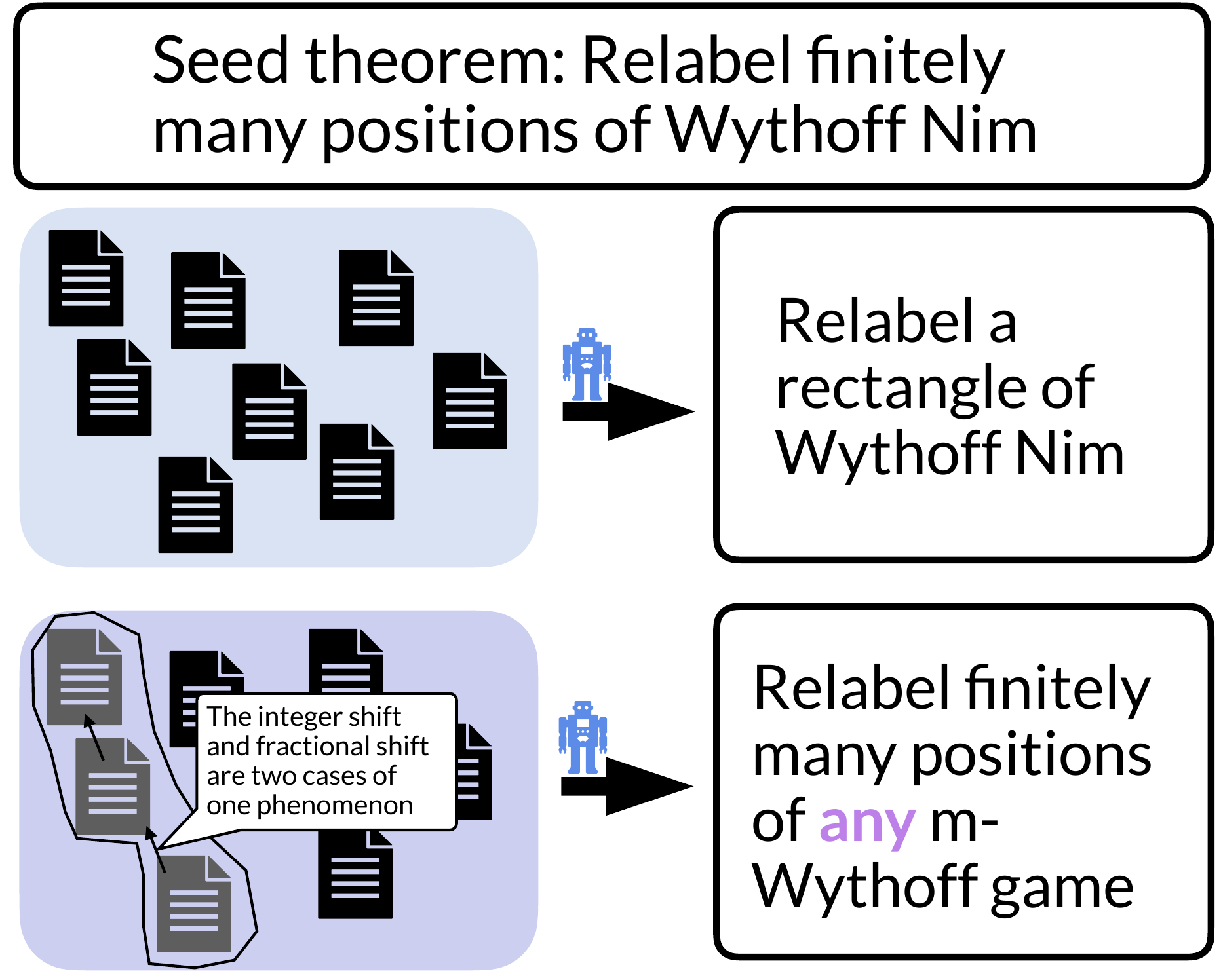}
\caption{
Two agents given the same seed theorem on relabeling finitely many positions of Wythoff Nim \citep{hu2024wythoff}.
The unconstrained agent (top) proposes a conjecture within the seed's own setting: relabeling a rectangle of Wythoff Nim.
The TEG agent (bottom) proposes a broader conjecture that unifies the seed theorem's integer shift and Larsson's fractional shift for $m$-Wythoff Nim \citep{larsson2009restrictions}: the phenomenon holds for any finite relabeling of any $m$-Wythoff game.
}
\label{fig:teaser}
\end{figure}

This paper contributes:
\begin{itemize}
    \item The Theorem Evolution Graph (TEG), a representation of how mathematical theorems build on one another.
    \item A method for LLM agents to construct and use TEGs for conjecture generation.
    \item Empirical evidence of TEG's utility: LLM and expert judges prefer TEG-generated conjectures, and ablations suggest the positive effects of paper selection, TEG construction, and graph structure.
\end{itemize}

\section{Related work}

\paragraph{LLMs for conjecture generation.}
Work on mathematical conjecture generation uses counterexample search to filter proposed statements~\citep{chuharski2024mining}, generates formal conjectures from existing mathematical libraries~\citep{onda2025leanconjecturer}, and jointly trains conjecturing and proving systems~\citep{dong2025stp,Poesia2024LearningFMA}. Other systems address problem selection: \citet{zheng2026problem} retrieve open problems from the mathematical literature and prioritize candidates for model attempts and expert review. Our work complements these approaches by studying how encoding the evolutionary differences between theorems can guide the generation of new conjectures.

\paragraph{Graph representations of scientific knowledge.}
Knowledge graphs link entities with relations~\citep{hogan2021knowledge}, with temporal variants also including when those relations hold~\citep{cai2023temporal}. Literature graphs connect papers, authors, and concepts~\citep{ammar2018semanticscholar}, which inspire citation-based methods that support document representation~\citep{cohan2020specter} and identify substantive uses of prior work~\citep{valenzuela2015identifying}. In mathematics, Prove2Me organizes theorem dependencies for proof construction~\citep{chen2026prove2me}. TEGs instead encode how theorem statements evolve: edges describe changes in hypotheses, conclusions, and settings, rather than citations or proof dependencies.

\paragraph{Literature-grounded research ideation.}
Prior work grounds LLM-based research ideation in literature to retrieve inspirations~\citep{wang2024scimon}, connect scientific concepts~\citep{baek2025researchagent}, assess novelty~\citep{yamada2025aiscientistv2}, and generate hypotheses and theories~\citep{vasu2025hyper,jansen2026scientific}. Other methods explicitly represent research progression in the literature: Chain of Ideas organizes papers into developmental chains~\citep{li2025chain}, while Graphs of Research uses temporally ordered citation graphs as supervision for idea generation~\citep{gao2026graphs}. TEGs distill papers into their core mathematical claims, represented as individual theorem nodes, and make mathematical change itself the organizing principle. The graph encodes not just which work builds on which, but what the mathematical advance consists of, giving agents a basis for proposing the next advance.

\section{Theorem Evolution Graphs}
\label{sec:teg}

Here we detail how to construct a Theorem Evolution Graph (TEG). Given a seed paper, an LLM agent constructs a TEG by \emph{identifying} relevant literature through backward reference (papers the seed paper cites) and forward citation (papers that cite the seed paper) traversal and \emph{comparing} the hypotheses, conclusions, and settings of its theorems to record how one extends or refines another. The agent uses these comparisons to propose next steps, updating the graph and its candidate conjectures.

\subsection{Graph Representation}
\label{sec:teg-representation}

A TEG is a directed graph $G = (V, E)$ where nodes represent individual theorems and edges describe mathematical changes between them.

\paragraph{Theorem nodes.}
Each node $v \in V$ is a theorem, extracted by the agent from the paper's text and denoted as $v = (\sigma_v, H_v, \kappa_v, s_v)$: its mathematical setting $\sigma_v$ (e.g., finite groups), hypotheses $H_v$, conclusion $\kappa_v$, and source $s_v$ (paper and theorem number). This representation isolates the components that may change between related theorems. These changes are subsequently encoded in edges.

\paragraph{Evolution edges.}
An edge $(u \to v) \in E$ records how theorem $v$ differs from, and builds on, theorem $u$.
Each edge carries an annotation generated by the agent,
$\delta_{uv} = (\Delta_{uv}, \iota_{uv})$.
The change description $\Delta_{uv}$ compares the settings, hypotheses, and conclusions of the two theorems, recording the change from $u$ to $v$.
The annotation $\iota_{uv}$ additionally records the new idea needed to make the step from one theorem to another.

\subsection{Constructing a TEG}
\label{sec:teg-construction}

TEG construction involves two operations: \emph{identifying} relevant theorems and \emph{comparing} them with those already represented. 
% The agent maintains the graph in a Markdown file \texttt{teg.md}.

\paragraph{Identify.}
\label{sec:teg-select}
The agent identifies relevant papers through Semantic Scholar\footnote{We use the Academic Graph API endpoints at \url{https://api.semanticscholar.org/api-docs/graph}.} endpoints that list a paper's references (backward search) and the papers that cite it (forward search). It follows references backward to find earlier theorems on which those in the graph build, and citations forward to find later theorems that extend or refine them. The agent then calls \texttt{read\_paper}, which processes a paper's full text and returns its extracted abstract, introduction, and theorem statements.

From each paper, the agent identifies relevant theorems and represents them in the canonical form defined in Section~\ref{sec:teg-representation}: $v = (\sigma_v, H_v, \kappa_v, s_v)$, specifying the setting, hypotheses, conclusion, and source paper and theorem number. This gives the agent a common representation for comparing theorems across papers.

\paragraph{Compare.}
\label{sec:teg-compare}

The agent reads the existing nodes and determines how the nodes are related. It selects comparison pairs using the citations between papers and statements in the paper, such as a claim to generalize an earlier theorem. When theorem $v$ is classified as building on theorem $u$, it records an edge $u \to v$.

For each edge, the agent records the changes from $u$ to $v$ in setting, hypotheses, and conclusion. It also generates a description of the mathematical idea introduced, based on the two theorem nodes and the changes between them.

\subsection{Using a TEG to generate conjectures}
\label{sec:teg-conjecturing}

The agent constructs the TEG starting with the seed paper’s main theorems and alternating identify and compare operations as it reads additional papers. After reading a new paper, it records the relevant theorem nodes and edges connecting them to nodes already in the graph. If no relevant connections are found, it marks the paper as unused.

Given a seed paper $p_0$, the agent builds a TEG $G = (V,E)$ and a set of candidate conjectures $\mathcal{C}$ in an interleaved loop. It begins by adding the seed paper's main theorems to $V$, then traverses backward to the earlier results they build on and forward to later work that builds on them, adding each theorem to $V$ and each change between theorems to $E$. When a chain or pattern in $G$ suggests a next step, the agent adds a candidate $c$ to $\mathcal{C}$ and links it to the nodes $V_c \subseteq V$ and edges $E_c \subseteq E$ that motivated it. It then checks $c$ against $V$ and expands $G$ through further search to determine whether $c$ is already known. If it finds an existing theorem $v$ matching $c$, it updates $V \leftarrow V \cup \{v\}$ and $\mathcal{C} \leftarrow \mathcal{C} \setminus \{c\}$, adds the relevant edges, and continues from $v$; it also removes any $c$ already posed as a conjecture or open problem. Updates to $G$ thus suggest or refine candidates, and candidates direct further expansion of $G$. It then selects one final conjecture $c^\star \in \mathcal{C}$.

\section{Experimental Setup}
\label{sec:setup}

We evaluate whether and how TEGs improve conjecture generation in three experiments: (1) TEG agents against an unconstrained baseline; (2) a crossed design that separates the papers an agent reads from whether it creates a TEG to guide its generation of conjectures; and (3) ablations on graphs that remove nodes or randomize edges.

\subsection{Test set}
\label{sec:setup-seeds}

We evaluate on 30 seed papers from two sources.
We select 18 papers from Math-Graph~\cite{alexander2026semanticsearch9million} that satisfy three criteria:
(1) the full source text is available;
(2) the paper proves at least one original theorem (i.e. not a survey paper); and
(3) the subject matter is accessible with a graduate-level background in computer science or discrete mathematics (allowing the authors to check results and analyze the data).
The remaining 12 papers are authored by a collaborating expert mathematician, who selected papers whose generated follow-up conjectures they felt confident assessing. Appendix~\ref{app:seeds} lists the seed papers.

\subsection{Agent setup}
\label{sec:setup-agents}

We use Claude Opus 5.5 and the OpenHands SDK and log all agent steps and reasoning. Within each experiment, conditions share the generation model, budget, task instructions, and output requirements. Access to literature and instructions for constructing or using the graph vary by experiment. Details are in
Appendices~\ref{app:agent-prompts}
and~\ref{app:evaluation-prompts}.

\paragraph{Tools and budget.}
Agents have tools for paper search, reading, reference lookup, and citation lookup, using arXiv, OpenAlex, and Semantic Scholar.
They also have a sandboxed terminal without internet access so the agent could run computations. Each run has a budget of \$7, 150 steps, and 60 minutes, with a warning at 85\% of the budget. Experiments~2 and~3 restrict literature access to a fixed collection of papers.

\paragraph{Task and output.}
Agents are instructed to generate one new conjecture from a seed paper. A new conjecture is defined as one not already articulated in the literature or posed as a conjecture or open problem in the literature. Agents generated the conjectures in a bulleted description of the setting, and a statement of at most 300 words. This standardization was intended to reduce differences in the conjecture format across conditions.

\subsection{Experiment Design}
\label{sec:setup-design}

\subsubsection{TEG vs.\ Unconstrained Search}
\label{sec:setup-exp1}

For each seed paper, a TEG agent and an agent without any constraint on how its conjecture generation strategy (referred to as an unconstrained agent) each generate one conjecture, yielding 60 conjectures. Both use the same model, tools, budget, and task instructions. The TEG agent constructs and uses a TEG following Section~\ref{sec:teg}, while the unconstrained agent chooses its own search and note-taking strategy. This comparison evaluates the effects of TEG-based conjecture generation compared to unconstrained conjecture generation.

\subsubsection{Paper Selection and TEG Construction}
\label{sec:setup-exp2}

To understand the effects of paper selection and theorem comparison, corresponding to the identify and compare processes in TEG construction, we give new agents fixed paper sets from the experiment in Section \ref{sec:setup-exp1}. Each set contains the seed paper. One set contains all papers opened by the TEG agent; the other contains all papers opened by the unconstrained agent. Agents can read papers in their assigned set but cannot retrieve additional literature.

We evaluate the following conditions on the test set: (1) papers opened by the TEG agent, with instructions to construct a TEG; (2) the same papers, with an unconstrained agent; and (3) papers opened by the unconstrained agent, with instructions to construct a TEG for conjecture generation.

Comparing conditions (1) and (2) tests the TEG construction and comparison while holding the available papers fixed. Comparing conditions (1) and (3) tests paper selection while holding the TEG procedure fixed. All conditions use the same model, budget, and output requirements. We report comparisons for seeds with completed outputs in both conditions: 27 seeds for the paper-selection comparison and 28 for each construction comparison (for a small number of papers, agents timed-out during the conjecture generation task which we removed from analysis).

\subsubsection{Graph Structure}
\label{sec:setup-exp3}

To understand which components of the TEG support conjecture generation, we test agents variants of the TEGs from the experiment in Section \ref{sec:setup-exp1}. For each seed paper, all conditions get the same paper set used by the original TEG agent. Agents can read these papers but cannot retrieve additional literature or modify the supplied graph.

Table~\ref{tab:exp3-conditions} defines the six variants. All variants use the same theorem nodes except seed-only, which uses only the seed paper's theorems. We evaluate all six variants on nine seed papers in the test set (chosen at random).

\begin{table}[t]
\centering
\small
\begin{tabular}{@{}lp{0.72\linewidth}@{}}
\hline
\textbf{Variant} & \textbf{Graph supplied to the agent} \\
\hline
Full & Full graph, including edge descriptions. \\
Unannotated & Same nodes and directed edges, without edge descriptions. \\
Shuffled & Same nodes and number of edges, with edges placed between random node pairs and no descriptions. \\
Similarity & Same nodes, linked by textual similarity using TF-IDF, with at most two links per node. \\
Edgeless & Same nodes, with all edges removed. \\
Seed-only & Only the seed paper's theorem nodes and the described edges among them. \\
\hline
\end{tabular}
\caption{Graph variants used in Experiment~3. Within each seed, agents receive the same papers and cannot modify the supplied graph.}
\label{tab:exp3-conditions}
\end{table}

\subsection{Evaluation}
\label{sec:setup-eval}

We evaluate conjectures in two stages: an individual assessment followed by a pairwise comparison. 
% We also qualitatively analyze agent behavior and compare LLM preferences with expert judgments.

\paragraph{Individual assessment.}
A referee agent evaluates the final conjecture. It is asked whether a strong researcher in the area should spend substantial time on the conjecture, a question close to the intellectual-merit criterion of NSF proposal review~\citep{nsf-pappg}. It grades the conjecture on clarity, correctness, novelty, difficulty, and significance, and raises objections classified as blocking, major, or minor. A different model independently checks each blocking or major objection. The verdict is fail if a blocking objection is upheld, revise if a major objection is upheld and no blocking objection is, and pass otherwise (analogous to a journal's reject, revise, and accept decisions).

The experiment in Section~\ref{sec:setup-exp1} uses three judge models: GPT-6.1 Sol, Grok 4.7, and Claude Sonnet 5.5. After observing similar performance across these models in pilot experiments of 10 papers, we use GPT-6.1 Sol as the sole referee in Sections~\ref{sec:setup-exp2} and~\ref{sec:setup-exp3}, with Grok and Sonnet checking its objections.

\paragraph{Pairwise preference.}
We use pairwise comparisons to assess which conjectures are worth pursuing, in line with prior work that states that comparative evaluation can better align LLM judgments with human assessments than absolute scoring~\citep{liusie2024comparative,liu2024aligning}. Experiments use GPT-6.1 Sol, Grok 4.7, and Claude Sonnet 5.5
as pairwise judges. Judges receive two conjectures generated from the same seed, together with the findings from the individual assessments, and choose which they would choose based on its potential contribution to the field. This reflects the emphasis on advancing knowledge in NSF's intellectual merit criterion~\citep{nsfMeritReview}. To further characterize these preferences, judges also compare significance, novelty, plausibility, clarity, ambition, and insight. This follows idea-quality metrics proposed in prior research-ideation work~\citep{si2025novel,li2025chain}.

Each judge evaluates each pair twice, with the conjectures presented in opposite orders. For the overall preference, we record a preference if the judge chooses the same conjecture both times; else, we record a tie. For individual criteria, where judges may also answer that the two conjectures are tied, we count the presentation orders separately.

\paragraph{Statistical analysis.}
We aggregate preferences across judges and seeds and report wins, losses, and ties. We calculate preference shares among non-tied judgments, with 95\% Wilson confidence intervals and two-sided binomial tests.

% \paragraph{Expert review.}
% A collaborating mathematician, an author of the 12 seed papers, compares the TEG and unconstrained conjectures for each paper, blind to condition.

% \paragraph{Agent behavior.}
% We analyze the papers agents read and open, their notes and graphs, and how they allocate actions across reading, computation, and conjecture development to connect these behaviors to differences in the conjectures produced.

\section{Results}

\subsection{Conjecture quality}
\label{sec:rq1}

\textbf{Judge preferences.} Judges preferred conjectures generated using a TEG over those generated by the unconstrained agent (Table~\ref{tab:rq1-preference}). Across 90 judge–paper pairs (three judges, 30 papers), judges preferred the TEG conjecture in 50 (56\%), the baseline conjecture in 29 (32\%), and judged the two tied in 11 (12\%). Among non-tied judgments, TEG conjectures were preferred in 63\% of cases (95\% CI [52\%, 73\%]; two-sided binomial test, $p = 0.02$), and each judge individually preferred TEG more often (17--9, 17--10, and 16--10).\footnote{The judgment-level confidence intervals and tests assume independence across judgments. Assessments of the same conjecture pair may be correlated across judges and presentation orders. In a separate analysis treating each seed paper as one unit, the TEG preference share is 63\% (95\% CI [47\%, 79\%]; $p = 0.13$); $p = 0.07$ for both the significance and plausibility criteria in the paper-level tests.}

Conjectures generated with a TEG were preferred for their significance, ambition, and insight, while baseline conjectures were preferred for their plausibility. This suggests a trade-off: TEGs help agents propose conjectures with greater perceived mathematical value, but whose plausibility might be less certain.

% Preamble: \usepackage{booktabs}

\begin{table*}[t]
\centering
\small
\setlength{\tabcolsep}{12pt}
\renewcommand{\arraystretch}{1.15}
\begin{tabular}{@{}lrrrcl@{}}
\toprule
Criterion & TEG & Baseline & TEG preferred (\%) & 95\% CI & $p$ \\
\midrule
Overall preference & 50 & 29 & 63 & [52, 73] & $0.02^{*}$ \\
\midrule
Significance & 103 & 54 & 66 & [58, 73] & $<0.001^{*}$ \\
Novelty      &  59 & 43 & 58 & [48, 67] & $0.14$ \\
Plausibility &  39 & 84 & 32 & [24, 40] & $<0.001^{*}$ \\
Clarity      &  16 & 19 & 46 & [30, 62] & $0.74$ \\
Ambition     &  90 & 50 & 64 & [56, 72] & $<0.001^{*}$ \\
Insight      &  96 & 57 & 63 & [55, 70] & $0.002^{*}$ \\
\bottomrule
\end{tabular}
\caption{
Preferences for the 30 seed papers.
The overall preference counts a judge's choice only if it holds in both presentation orders (90 judgments; remaining judgments are ties); criteria count the two presentation orders separately (180 judgments).
Percentages exclude ties, with values below 50\% favoring the baseline.
Intervals are 95\% Wilson confidence intervals.
$^{*}p<0.05$ in a two-sided binomial test against 50\%,
uncorrected for multiple comparisons.
}
\label{tab:rq1-preference}
\end{table*}

\textbf{Expert review.} An expert mathematician, an author of the 12 seed papers reviewed, compared the TEG and baseline conjectures for each paper without knowing which condition produced them. The expert preferred the TEG conjecture on 8 papers and the baseline conjecture on 4, with no ties: 67\% TEG preference (95\% CI [39\%, 86\%]), close to the panel's 63\%.

The expert's preferences agreed with the LLM panel's majority preference on 9 of 12 papers (Cohen's $\kappa = 0.47$; Krippendorff's $\alpha = 0.49$).

% Once complete, add an "Expert preference" row to
% Table~\ref{tab:rq1-preference}, separated from the LLM judgments.
% Keep individual expert comments and per-seed outcomes in the appendix.

\textbf{Agent behavior.} Agents in the two conditions read different papers. Counting the papers each agent opened (excluding the seed paper), the two agents opened 149 papers across the 30 seeds, of which only 44 (30\%) were opened by both.

\begin{figure}[t]
\centering
\includegraphics[width=\linewidth]{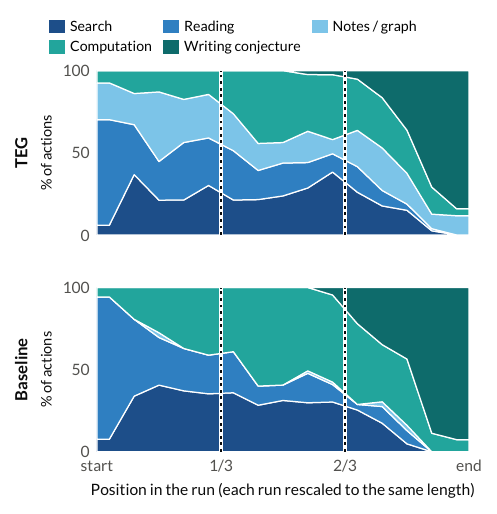}
\caption{
Agent action allocation over the run. Each run is rescaled to the same length and divided into 15 equal bins by action count; shares pool the actions of all 30 runs in a condition, and dotted lines mark the thirds. Baseline agents use computation earlier, while TEG agents devote early actions to building the graph.
}
\label{fig:rq1-actions}
\end{figure}

The agents also differed in how they used their tools (Figure~\ref{fig:rq1-actions}). Baseline agents used computation earlier: computation accounted for 27\% of their actions in the first third of a run, compared with 13\% for TEG agents. Instead, TEG agents spent 28\% of their early actions building the graph; both conditions spent a similar share of early actions reading papers (35\% and 40\%). Baseline agents also committed to a conjecture slightly sooner: on average, they first wrote a draft 82\% of the way through a run, compared with 87\% for TEG agents (TEG wrote later on 23 of 30 seeds; mean paired difference 4.4 percentage points, 95\% CI [1.5, 7.2]).

\textbf{Case study.} To concretize these results, we analyze how two agents given \citet{hu2024wythoff} as a seed generate new conjectures. In Wythoff Nim, two players alternately remove tokens from two piles, taking any positive number from one pile or the same number from both, and the player who takes the last token wins; the losing positions are exactly the pairs $(\lfloor n\phi \rfloor, \lfloor n\phi^2 \rfloor)$ and their reflections, where $\phi$ is the golden ratio. The paper proves that after finitely many positions of Wythoff Nim are relabeled as winning or losing, the losing positions agree asymptotically with the original pattern after an integer translation.

Both agents first asked whether this extends to $m$-Wythoff Nim, a family of related games, and both found that integer translations failed to match the simulated losing positions. The baseline agent treated this as a boundary (``the defect structure is specific to Wythoff's Nim'') and returned to ordinary Wythoff Nim. It tested 406 rectangular alterations and proposed a refinement of the seed's result for this setting.

The TEG agent recorded the failed candidate and investigated further. Following references, it found \citet{larsson2009restrictions}, two citation steps from the seed, through \citet{larsson2010generalized}. Larsson establishes this behavior for a special family of alterations of $m$-Wythoff. The agent recorded the contrast between the two theorems: the seed covers all finite alterations of ordinary Wythoff Nim with an integer translation, while Larsson covers a restricted family of alterations of $m$-Wythoff with a fractional phase. The agent proposed a formula for the phases and checked that it reproduces Larsson's results and holds for sampled random alterations. Its final conjecture unifies the two theorems: every finite alteration of every $m$-Wythoff game yields Beatty sequences with computed phases, up to $O(\log K)$ exceptions among the first $K$ positions. The LLM judges preferred the TEG conjecture.

\subsection{Effects of paper selection and TEG construction}
\label{sec:rq2}

We isolate the effects of paper selection and TEG construction to conjecture generation. Specifically, does the TEG's advantage come from the papers agents find by following citations backward and forward from the seed?

\begin{figure*}[t]
\centering
\includegraphics[width=\textwidth]{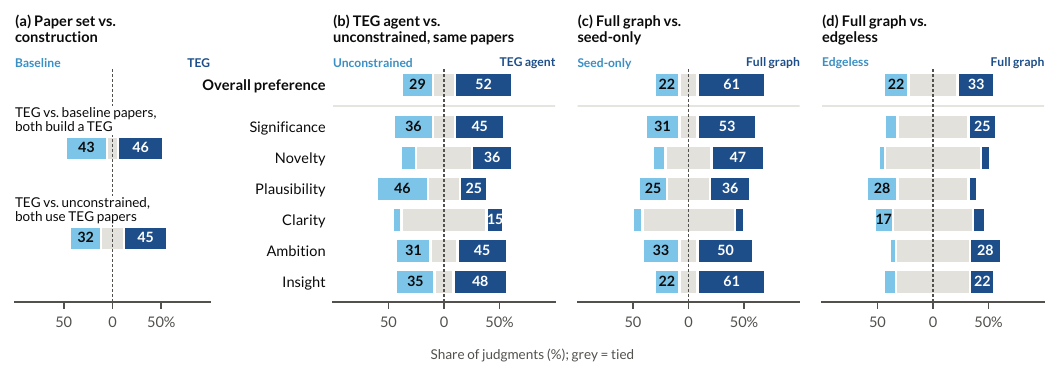}
\caption{
Effects of paper selection, TEG construction, and graph components.
(a) Overall preferences when varying the paper set (both agents build a TEG) or the construction method (both use TEG-selected papers).
(b) The original TEG agent compared with an unconstrained agent given its papers.
(c, d) Conjectures from the full graph compared with those from seed-only and edgeless graphs.
Bars show the share of judgments favoring the TEG side (right), tied (grey), or the alternative (left). Overall preferences count a judge's choice only if it holds in both presentation orders; criteria count the two orders separately.
}
\label{fig:rq2-pairwise}\label{fig:rq3-pairwise}
\end{figure*}

\textbf{Paper selection.} When both agents construct a TEG, judges show no clear preference between the two paper sets. Across 54 judge--seed comparisons, they prefer conjectures based on TEG-selected papers in 46\% of cases and baseline-selected papers in 43\%, with 11\% tied. Among non-tied judgments, 52\% favor TEG-selected papers (95\% Wilson CI [38\%, 66\%]; $p = 0.89$; Figure~\ref{fig:rq2-pairwise}a). This comparison provides no evidence that TEG-selected papers alone improve conjecture quality.

\textbf{TEG Construction.} The preference for TEG-enabled conjectures holds when paper selection is fixed. Given the same TEG-selected papers, judges prefer conjectures from an agent building a TEG over those from an unconstrained agent in 25 of 56 comparisons and the unconstrained agent's in 18, with 13 ties (58\% of non-tied judgments; 95\% CI [43\%, 72\%]). The advantage holds against the original TEG agent, which judges prefer over the unconstrained agent given its papers in 29 of 56 comparisons, against 16, with 11 ties (64\% [50\%, 77\%]; Figure~\ref{fig:rq2-pairwise}a). This matches the 63\% preference in Section~\ref{sec:setup-exp1}. Although the overall preferences do not reach statistical significance at this sample size (p = 0.36 and 0.07), both comparisons favor TEG construction. This pattern is consistent with a benefit from constructing and using the graph, beyond the choice of papers alone.

The criterion-level judgments mirror the trade-off in Section~\ref{sec:setup-exp1}. Compared with the original TEG agent, the unconstrained agent given its papers produces conjectures judged less novel (15 vs.\ 40 judgments; 73\% favor TEG, 95\% CI [60\%, 83\%]; $p = 0.001$) but more plausible (52 vs.\ 28; 65\% favor the unconstrained agent, 95\% CI [54\%, 75\%]; $p = 0.01$; Figure~\ref{fig:rq2-pairwise}b).

\subsection{Ablations}
\label{sec:rq3}

To observe the contribution of the literature beyond the seed paper, we compare the full graph with a graph containing only the seed paper's theorems. Judges prefer full-graph conjectures in 11 of 18 comparisons and seed-only conjectures in 4, with 3 ties. Among non-tied judgments, 73\% favor the full graph (95\% Wilson CI [48\%, 89\%]; $p = 0.12$; Figure~\ref{fig:rq3-pairwise}c). Full-graph conjectures are judged more novel (17 vs.\ 4 preferences; 81\% [60\%, 92\%]; $p = 0.007$) and more insightful (22 vs.\ 8; 73\% [56\%, 86\%]; $p = 0.02$), without a loss in plausibility (13 vs.\ 9; 59\% [39\%, 77\%]).

We also compare the full graph with a graph with the theorem nodes but no edges. Judges prefer full-graph conjectures in 6 of 18 comparisons and edgeless-graph conjectures in 4, with 8 ties. Among non-tied judgments, 60\% favor the full graph (95\% Wilson CI [31\%, 83\%]; $p = 0.75$; Figure~\ref{fig:rq3-pairwise}d). 

Conjectures generated with edges are judged more ambitious (10 vs.\ 2 preferences; 83\% [55\%, 95\%]; $p = 0.04$; Figure~\ref{fig:rq3-pairwise}d). Judges also favor them for significance (9 vs.\ 4; 69\% [42\%, 87\%]; $p = 0.27$), while favoring edgeless-graph conjectures for plausibility (10 vs.\ 3; 77\% [50\%, 92\%]; $p = 0.09$), although neither difference is statistically significant. This pattern resembles the trade-off between perceived mathematical value and plausibility observed in Section~\ref{sec:rq1}.

Edge descriptions also appear to help agents avoid false conjectures. The referee agent flagged none of the nine conjectures generated in the Full condition as false, compared with four of the 18 conjectures generated across the Unannotated and Shuffled conditions combined. This is consistent with a benefit from annotated theorem relationships, although pooling Unannotated and Shuffled does not isolate the effect of edge descriptions because Shuffled also changes which nodes are connected.

\section{Discussion}
\label{sec:discussion}

Our results suggest that TEGs help agents propose conjectures with greater significance, ambition, and insight, with plausibility graded as lower. The ablations suggest that how agents compare theorems may matter more than which papers they select. Qualitative examples offer a possible explanation: recording changes between theorems can help agents identify a step beyond the seed paper's setting, and understanding whether the resulting conjecture is plausible can be more difficult.

\paragraph{Qualitative examples.}
The examples illustrate how theorem comparison can help agents identify new directions and sharpen their claims. In one case, the TEG agent traced a sequence of improvements and proposed a new way to generalize them, while the ordinary-notes agent proposed a narrower claim that largely followed from work already in its library. In another, both agents reached the same broad idea, but the TEG agent recognized a prior theorem as a special case its conjecture should recover. This gave it a concrete check that helped turn a tentative pattern into a more precise, testable claim. These cases suggest that TEGs help agents use prior work both to propose the next advance and to specify what it must explain.

\paragraph{Future work.}
These findings motivate treating conjecture generation as structured search over mathematical moves. A system could first choose how to extend a theorem, such as by weakening a hypothesis, strengthening a conclusion, or changing the setting, and then instantiate that move as a conjecture. These could be actions in search or reinforcement learning, with feedback from counterexample search, proof attempts, novelty checks, and expert preferences. A central challenge is balancing these signals: rewarding provability alone may favor routine consequences, while rewarding ambition alone may favor implausible claims. TEGs provide a representation in which these choices can be made explicit and studied.

TEGs could also support workflows in which mathematicians request particular kinds of extensions and inspect the theorems motivating each candidate. A mathematician could correct an inferred relationship, identify a missing hypothesis, or suggest a different setting to explore. The agent could then update the graph and revise its candidates. Evaluating whether this interaction helps mathematicians identify worthwhile questions would provide a stronger test of practical utility than preference judgments alone.

\section{Limitations}
Our study covers a small selected set of seed papers and one generation model because of budget restrictions, so findings may not generalize across mathematical fields or models. We do not validate theorem extraction or edge annotations with experts at scale; errors in either can propagate into conjectures. Our evaluation measures assessed quality and preference rather than eventual mathematical value. TEGs also emphasize extensions of known theorems, and citation traversal favors developments connected through the literature. This may underrepresent conjectures with unexpected phenomena or connections between distant areas.

\section{Conclusion}
\label{sec:conclusion}

We introduced Theorem Evolution Graphs, which represent individual theorems and describe how their hypotheses, conclusions, and settings change. LLM agents construct and use these graphs to generate new conjectures. Judges preferred TEG-generated conjectures overall and favored their significance, ambition, and insight, while preferring baseline conjectures for plausibility. Ablations and case studies suggest that theorem comparisons help agents formulate broader and more precise extensions of prior work. TEGs offer a way to make these proposed mathematical steps inspectable, supporting further study of how agents can identify conjectures worth pursuing.

\section*{Ethical Considerations}
TEG agents use LLMs to read mathematical literature and propose conjectures. Although each conjecture builds on published theorems, it may be false, may misstate prior results, or may already be known. The referee agent rejected some generated conjectures as false or already in the literature, and none of the conjectures has been proved. They are candidates that need expert review before anyone invests research time in them. Agents were instructed to cite only papers they had opened and never to invent references, but these instructions do not guarantee correct attribution.

Our evaluation relies mainly on LLM referees and judges, which may favor particular writing styles or conjectures that sound ambitious. To reduce this risk, we use judges from three model families, present each pair in both orders, give judges objections that a second model had checked, and compare the panel's preferences with a blind expert review. That review covers 12 of the 30 seed papers, and its agreement with the panel is moderate ($\kappa = 0.47$), so it partly validates the LLM judgments but does not replace expert assessment. The expert participated voluntarily and was not compensated.

Automated conjecture generation could also make it easier to produce large numbers of unvetted conjectures or submissions, which would add to the load on reviewers and editors. TEGs are intended to support mathematicians in finding and critically assessing research directions, not to replace their judgment or the work of proving results. Because a TEG records the theorems and edges behind each conjecture, a reader can check where a proposed step comes from.

\paragraph{Use of AI assistants.}
We used Claude Code to implement the experiments and analyses, and GPT-5.6 to help draft and edit the manuscript. The authors reviewed all AI-assisted code and text.

\bibliography{custom}

\appendix

\section{Agent Instructions}
\label{app:agent-prompts}

Each agent's system prompt concatenates three layers: the task (Section~\ref{app:prompt-task}) and the environment (Section~\ref{app:prompt-env}), which are identical for both conditions, followed by a method layer that differs between them (Sections~\ref{app:prompt-teg} and~\ref{app:prompt-free}). Placeholders in double braces are filled in for each seed paper. The agent then receives the message ``Begin. The seed paper is arXiv:\{id\}.'' Each run has a budget of \$7, 150 steps, and 60 minutes; when 85\% of any of these is used, the agent receives the message ``Budget notice: you have used 85\% of your budget (money, steps, or time). Finish now: make sure conjecture.md is complete and self-contained, call check\_clarity on its final setting and statement if you have not already, then call finish.''

\subsection{Task (shared)}
\label{app:prompt-task}
\begin{promptbox}
# Task

Your goal is to formulate one new conjecture, starting from the seed paper, that an expert in this area would consider worth working on. A conjecture is worth working on when it is:

- **new**: not already proved, refuted, stated as a conjecture or open problem, or an immediate consequence of known results;
- **sound**: precisely stated and consistent with known examples and special cases;
- **interesting**: settling it would teach something, by connecting to or advancing known work.

## Output: `conjecture.md`

Write your conjecture in `conjecture.md`, which already contains these headings:

1. **Statement**
   - *Setting*: the standing assumptions, the objects involved, and every definition or notation that is not standard in the field.
   - *Conjecture*: one precise assertion, with all quantifiers and hypotheses explicit.

   The statement must be self-contained: an expert who has not read the seed paper must be able to understand it completely. Do not refer to the seed paper's numbered results, sections, or notation.
2. **The case for it**
   - *Relation to known results*: what it extends, refines, or contrasts with, and the searches you ran to check that it is new.
   - *Evidence*: the examples and special cases you checked it against, and what you found.
   - *Why it is worth working on*: what would follow if it were true, a possible route to a proof, and the main obstacle.
3. **References**: only the seed paper and papers you opened with `read_paper`, one per line, as: authors, *title*, arXiv id (or venue and year).

## Standard form

Every submission uses the same form:

- Write all mathematics in LaTeX: `$...$` inline and `$$...$$` for displays. Do not use Unicode math symbols (such as (*$\leq$*), (*$\in$*), (*$\theta$*), (*$\mathbb{R}$*), (*$\mathcal{F}$*)).
- *Setting* is a bulleted list. Each bullet introduces one object, definition, or standing assumption. Define every notation that is not standard in the field, and nothing that is.
- *Conjecture* is one paragraph stating a single assertion. You may follow it with one sentence beginning "In words:".
- The Statement (Setting and Conjecture together) has at most 300 words. Examples, remarks, and motivation belong in the case for it, not in the Statement.
- The case for it has at most 1500 words.
\end{promptbox}

\subsection{Environment (shared)}
\label{app:prompt-env}
\begin{promptbox}
# Environment

You are working alone on a mathematics research task, and nobody will answer questions. Make reasonable judgments and keep going until you have written `conjecture.md` and called `finish`.

Your working directory contains:

- `paper/paper.tex`: the full LaTeX source of the seed paper, "{{title}}" by {{authors}} (arXiv:{{arxiv_id}}). You can also read it in layers with `read_paper`;
- `paper/metadata.json`: its title, authors, abstract, and categories;
- `notes/`: your working notes;
- `conjecture.md`: the file you will write your conjecture in.

## Tools

- `search_papers(query)`: searches the literature and returns matching papers with their ids.
- `read_paper(id, part)`: returns a paper's text in layers. `"overview"` (the default) gives the abstract, section list, introduction, and main statements; `"section N"` gives one section; `"full"` gives the whole paper.
- `get_references(id)`: lists the papers a paper cites, with the sentences in which they are cited.
- `get_citations(id)`: lists the papers that cite it, with the sentences in which they cite it.
- `check_clarity(setting, statement)`: a reader who sees nothing but this text reports any clarity problems in it.
- `file_editor`: read and write files in your working directory.
- `sandbox_terminal`: run commands for computations (Python with SymPy, NumPy, mpmath). It has no internet access.

Use only the paper tools to access the literature.

## Budget

You have ${{budget_usd}}, {{budget_steps}} steps, and {{budget_minutes}} minutes of wall-clock time; each sandbox command counts toward the time while it runs. When 85% of any of these has been used, you will receive a message asking you to finish. If the budget runs out before `conjecture.md` is written, the run counts as no submission.

# Research rules

- Cite only the seed paper and papers you opened with `read_paper`. Never invent references, theorem numbers, or quotations. If you rely on something from memory, say so.
- Before finalizing, search the literature for your statement and close variants of it.
- Check your statement against concrete examples and known special cases.
- Label your evidence honestly: `[Verified]` for a check over a stated range, `[Heuristic]` for a suggestive argument. Never describe the conjecture as proved.
- `conjecture.md` must be self-contained and in the standard form described above. It must not mention your notes, your working files, or how you worked.
- Before finishing, call `check_clarity` on your final setting and statement at least once, and fix the problems it reports that you agree with.
\end{promptbox}

\subsection{Method: TEG agent}
\label{app:prompt-teg}
The TEG agent's working directory also contains \texttt{teg.md},
pre-filled with the four section headings below.
The recording rule is enforced by \texttt{read\_paper}: until the
last paper opened is mentioned in \texttt{teg.md}, a request to
open a new paper is refused. The agent must record that paper
as the source of a theorem or edge, or under ``Read, not used''
with a one-line reason. Previously opened papers can be reread
at any time.

\begin{promptbox}
# Working method: Theorem Evolution Graph

As you read, build a Theorem Evolution Graph in `teg.md`. Use it to decide what to read next and what to conjecture.

**Literature access.** Discover papers only by following references backward or citations forward from the seed paper or papers reached through this traversal.

**Starting order** (return to any step whenever it helps):

1. **Anchor.** Record the main theorems of the seed paper as nodes.
2. **Backward.** For each main theorem, identify earlier theorems it builds on. Start with the paper's descriptions of prior work and follow its references. Continue through references of papers you reach when relevant.
3. **Forward.** Follow citing papers to identify later theorems that extend or refine those in the graph. Continue through citations of papers you reach when relevant.
4. **Throughout, reason over the graph.** When a comparison, chain, or pattern suggests a next step, record a candidate conjecture and link it to the nodes and edges that motivated it. Use these candidates to guide further citation traversal. Check them against theorems already in the graph and additional papers reached through references or citations. If a candidate is an established theorem, add it to the graph with the relevant edges and explore further extensions. Exclude candidates already posed as conjectures or open problems.

**Identify and compare.** After reading each paper, select its relevant theorems and record their settings, hypotheses, conclusions, and sources. Compare them with related theorems already in the graph. In the same edit, record the new theorem nodes and the edges connecting them to existing nodes. Direct each edge from the earlier theorem to the theorem that extends or refines it, regardless of the order in which you read them.

For each edge, describe what changed in the setting, hypotheses, and conclusion. Record the mathematical idea needed for the change.

**Recording rule.** After reading a paper, record it in `teg.md` before opening another: either as the source of a node or edge, or under "Read, not used" with a reason. `read_paper` will not open a new paper until you do.

Within the available budget, select one candidate developed through your reasoning over the graph as your final conjecture.

**Format of `teg.md`** (the file already contains the section headings):

```
# Theorem Evolution Graph

## Theorems
### T1: <short name>
- Source: <paper id>, <theorem number> (if known only through another paper, add: via <that paper's id>)
- Setting: ...
- Hypotheses: ...
- Conclusion: ...

## Edges
### T1 (*$\rightarrow$*) T2 (T2 builds on T1)
- Change: <what changed from T1 to T2 in setting, hypotheses, and conclusion>
- New idea needed: <idea>
- Source: stated by the authors (<location>) | inferred by me

## Candidate conjectures
### C1: <short name> status: active | revised into C3 | known (now T7) | dropped (<reason>)
- Extends: <which nodes and edges, and the step taken>
- Statement: <...>
- Checked against: <theorem nodes and papers reached through citation traversal>

## Read, not used
- <paper id>: <one-line reason>
```
\end{promptbox}

\subsection{Method: unconstrained agent}
\label{app:prompt-free}
\begin{promptbox}
# Working method

Work however you judge best. You may keep notes in `notes/` in any form.
\end{promptbox}

\subsection{Variants for Experiments 2 and 3}
In Experiments~2 and~3, agents read from a fixed library of papers instead of searching. The environment layer replaces the four paper tools with a \texttt{read\_paper} tool restricted to the library, lists the library's papers, and asks the agent to check its statement against the library rather than search the literature; the task layer correspondingly asks how the agent checked novelty rather than which searches it ran. In Experiment~2, the TEG method layer asks the agent to find earlier and later results among the papers in its library. In Experiment~3, the method layer is replaced by the following, and the prepared graph is placed in \texttt{graph.md}:
\begin{promptbox}
# Working method: theorem graph

Your working directory also contains `graph.md`, a graph of theorems from the papers in your library, prepared before your run. Each theorem is listed with its source and, where available, its setting, hypotheses, and conclusion. If the graph has an `Edges` section, an arrow `A (*$\rightarrow$*) B` means that result B builds on result A; an edge may describe what changed. If it has a `Links` section, a link `A (*---*) B` means that results A and B are closely related.

Use the graph to decide what to read and what to conjecture. Do not edit `graph.md`. You may keep notes in `notes/` in any form.
\end{promptbox}

\section{Evaluation Instructions}
\label{app:evaluation-prompts}

\subsection{Models and budgets}
\label{app:models}
All models are served through Amazon Bedrock with high reasoning effort.

\begin{table}[h]
\centering
\small
\begin{tabular}{p{0.28\linewidth}p{0.36\linewidth}p{0.26\linewidth}}
\hline
Role & Model & Budget \\
\hline
Generating agents (TEG and baseline; all experiments) & Claude Opus 5.5 & \$7, 150 steps, 60 min per run \\
Clarity check tool & Claude Opus 5.5 & one call per use \\
Referees, Exp.~1 & GPT-6.1, Grok 4.7, and Claude Sonnet 5.5, one referee each & \$3.50, 60 steps per referee \\
Objection checks, Exp.~1 & Rotated across labs: Sonnet checks Grok, GPT checks Sonnet, Grok checks GPT & \$2, 40 steps per check \\
Referee, Exp.~2--3 & GPT-6.1 & \$3.50, 60 steps \\
Objection checks, Exp.~2--3 & Grok 4.7 (Exp.~2); Claude Sonnet 5.5 (Exp.~3) & \$2, 40 steps per check \\
Pairwise judges & Exp.~1: GPT-6.1, Grok 4.7, Claude Sonnet 5.5; Exp.~2--3: GPT-6.1, Claude Sonnet 5.5 & one call per order, no tools \\
\hline
\end{tabular}
\caption{Models and budgets. Referees and checkers are agents with the same paper tools and sandbox as the generating agents; an agent receives a request to finish once 85\% of any budget is used.}
\label{tab:models}
\end{table}

\subsection{Referee}
Each referee is an agent with the same paper tools and sandbox as the generating agents, a budget of \$3.50 and 60 steps, and the message ``Begin.''. Placeholders are filled with the seed paper and the conjecture's statement.
\begin{promptbox}
# Role

You are a referee. Your job is to decide whether a strong researcher in this area should spend substantial time on the research conjecture below. You work alone, and nobody will answer questions.

The conjecture was proposed starting from the paper "{{title}}" by {{authors}} (arXiv:{{arxiv_id}}). You can read that paper with `read_paper`.

**The burden of proof is on the conjecture.** It earns the disposition `proceed` only by surviving a serious attempt to find a defect: that it is false, already known, easy, or a routine variation that would teach little. Doubt that you could not resolve is a reason for `cannot_assess` or an objection, never for a favorable verdict.

# The conjecture

{{conjecture}}

# Referee brief

Assess the conjecture critically and independently. Look for counterexamples, hidden assumptions, published results that already contain it or imply it, and short arguments that settle it. Distinguish an actual defect from an unclear passage, an unverified citation, and a matter of taste. Do not repair the conjecture silently: if a change would help, record it as an objection and give the revision. Record what you checked, not only what you concluded.

# Criteria

Give a verdict on every criterion. **A favorable verdict must be backed by the evidence named below; without that evidence, use `cannot_assess`.**

1. **Clarity**: `clear` or `unclear`. Restate precisely what it asserts, and note any statement that can be read in two materially different ways.
2. **Correctness**: `plausible`, `wrong`, or `cannot_assess`.
   - `plausible` requires the specific checks you performed (special cases, small cases computed, consistency with known theorems) and what each showed.
   - `wrong` requires a counterexample or a contradiction with a known result.
3. **Novelty**: `new`, `already_done`, or `cannot_assess`. Search for the exact statement and for its key formula, threshold, or constant; look at later papers that cite the seed paper and its main references (`get_citations`), since a conjecture is often stated in a follow-up paper rather than in the original.
   - `new` requires the closest prior results you found (with their ids) and exactly how the conjecture goes beyond each.
   - `already_done` means the literature already contains it: proved, refuted, stated as known or as an open problem, or an immediate consequence of a published result. Cite where.
4. **Difficulty**: `not_easy`, `easy`, or `cannot_assess`.
   - `not_easy` requires your recorded attempt to settle it and the point where the attempt gets stuck.
   - `easy` requires a short argument using standard techniques; give the sketch.
5. **Significance**: `significant`, `routine`, `not_interesting`, or `cannot_assess`.
   - `significant` requires saying what would be learned from settling it that is not already understood (`what_would_be_learned` in the evidence).
   - `routine` means a variation of a known result whose answer is expected and would teach little beyond that result. A variation is not routine merely because of its form (for example, a weakened hypothesis or a new setting); say what makes it routine or significant.

# Objections

Record every objection. For each one give:

- `type`: `unclear`, `false`, `known`, `easy`, `routine`, `unsupported` (a favorable verdict you could not back with evidence), or `other`;
- `severity`:
  - `blocking`: it defeats the conjecture as stated;
  - `major`: it must be resolved before the conjecture is worth working on;
  - `minor`: it does not affect the decision;
- `location` (the part of the statement), `issue`, `consequence`, and `would_resolve` (what evidence or revision would settle it).

Each unfavorable verdict must have a matching objection of at least this severity:

| Verdict | Objection type | Severity |
|---|---|---|
| significance `routine` or `not_interesting` | `routine` | `blocking` |
| novelty `already_done` | `known` | `blocking` |
| correctness `wrong` | `false` | `blocking` |
| clarity `unclear` | `unclear` | `major` |
| difficulty `easy` | `easy` | `minor` (or higher, if you judge it so) |

An objection of type `false` (part of the statement is false, even in an edge case) is at least `major`.

# Disposition

The disposition must follow from your objections:

- `proceed`: no blocking or major objection.
- `revise`: at least one major objection, no blocking one, and a specific revision would make it worth working on. Give the revised statement.
- `stop`: at least one blocking objection.
- `cannot_assess`: you could not check enough to decide; say what is missing.

Explain why the disposition follows from the objections.

# Graded scores

If the conjecture is clear, score each from 1 to 3 (`null` if you cannot judge):

- `interest`: 1 = few would care; 2 = some experts in the area would care; 3 = many experts in the area would care.
- `advances_understanding`: 1 = settling it would add little; 2 = it would add something useful; 3 = it would significantly advance understanding of the area.
- `ambition`: 1 = a routine exercise; 2 = a substantial project; 3 = a major open problem.
- `new_mechanisms`: 1 = existing techniques clearly suffice; 2 = known techniques would need real adaptation; 3 = it needs fundamentally new ideas.

# Tools and budget

- `search_papers`, `read_paper`, `get_references`, `get_citations`: the literature.
- `sandbox_terminal`: computations (Python with SymPy, NumPy, mpmath). No internet access.
- `file_editor`: read and write files in your working directory.

You have ${{budget_usd}} and {{budget_steps}} steps. When 85% of the budget has been used, you will receive a message asking you to finish.

Cite only papers you opened with `read_paper`. Never invent references, theorem numbers, or quotations. If you rely on something from memory, say so.

# Output

Write two files in your working directory, then call `finish`. Your assessment counts only if it is in these files; text in a final message is not read.

1. `report.md` (its headings are already there): readable on its own by a mathematician.
2. `verdict.json`:

```json
{
  "clarity": "clear | unclear",
  "correctness": "plausible | wrong | cannot_assess",
  "novelty": "new | already_done | cannot_assess",
  "difficulty": "not_easy | easy | cannot_assess",
  "significance": "significant | routine | not_interesting | cannot_assess",
  "evidence": {
    "checks": "the checks behind your correctness verdict",
    "closest_prior_work": [{"id": "arXiv:...", "difference": "how the conjecture goes beyond it"}],
    "proof_attempt": "your attempt to settle it and where it gets stuck (or the sketch, if easy)",
    "what_would_be_learned": "what settling it would teach that is not already understood"
  },
  "objections": [
    {"id": "O1", "type": "known", "severity": "major", "location": "...", "issue": "...",
     "consequence": "...", "would_resolve": "..."}
  ],
  "disposition": "proceed | revise | stop | cannot_assess",
  "revision": "the revised statement, if the disposition is revise",
  "scores": {"interest": 1, "advances_understanding": 1, "ambition": 1, "new_mechanisms": 1}
}
```
\end{promptbox}

\subsection{Objection check}
Every blocking or major objection is checked by a second agent from a different model family, with a budget of \$2 and 40 steps.
\begin{promptbox}
# Role

You are an independent referee. Another referee raised the objection below against a research conjecture. Your only job is to decide whether that objection stands. You work alone, and nobody will answer questions.

The conjecture was proposed starting from the paper "{{title}}" by {{authors}} (arXiv:{{arxiv_id}}). You can read that paper with `read_paper`.

# The conjecture

{{conjecture}}

# The objection

{{objection}}

# Procedure

Check the objection's factual basis yourself; do not take it on trust.

- If it cites a result, open the source and check that the result says what the objection claims and applies under the conjecture's hypotheses.
- If it gives a counterexample, verify it.
- If it gives an argument or a proof sketch, check every step.
- If it says the conjecture is routine or unimportant, judge whether the stated reasons hold.

# Outcomes

- `standing`: the objection is correct, with the stated severity.
- `downgraded`: the objection is correct but less severe than stated; give the correct severity (`major` or `minor`).
- `resolved`: the objection is mistaken; say exactly why.
- `cannot_assess`: you could not check it; say what is missing.

# Tools and budget

You have the same tools as the original referee: `search_papers`, `read_paper`, `get_references`, `get_citations`, `sandbox_terminal`, and `file_editor`. You have ${{budget_usd}} and {{budget_steps}} steps. Cite only papers you opened with `read_paper`; if you rely on memory, say so.

# Output

Write `check.json` in your working directory, then call `finish`. Your assessment counts only if it is in this file.

```json
{"outcome": "standing | downgraded | resolved | cannot_assess",
 "severity": "blocking | major | minor",
 "reason": "what you checked and what you found"}
```

`severity` is the severity you find the objection to have (for `resolved`, use `minor`).
\end{promptbox}

\subsection{Pairwise preference}
Pairwise judges answer in a single model call, without tools. Each pair is presented in both orders. The verified findings for each conjecture list the blocking and major objections that stood after checking, each labelled by type (e.g., ``Already in the literature'') with the checker's confirmation; a conjecture with none receives ``No serious objection survived independent checking.''
\begin{promptbox}
# Role

You are an experienced research mathematician advising on how to spend research time.

Two conjectures, A and B, were proposed starting from the same paper: "{{title}}" by {{authors}} (arXiv:{{arxiv_id}}).

Independent referees have already examined each conjecture, with literature search and computation. Below each statement are their verified findings: objections that a second, independent referee checked and confirmed. Treat these findings as established facts. Beyond them, judge from your own knowledge.

# Conjecture A

{{statement_a}}

## Verified findings about A

{{findings_a}}

# Conjecture B

{{statement_b}}

## Verified findings about B

{{findings_b}}

# Question

Which of these two conjectures would contribute more to the research community if it were pursued? You must choose A or B.

These are proposals of **new** conjectures. A conjecture that already appears in the literature, whether proved, refuted, or posed as a conjecture or open problem, is not a new contribution. Do not prefer it because the problem itself is worth pursuing.

First, say whether A and B are substantively the same conjecture: equivalent statements, possibly in different notation or with different supporting material. Then make your choice either way.

# Consider

- **Significance**: how much would a resolution matter? Does it bear on questions the community cares about, or open the way to further results? Is it more than a routine variation of a known result?
- **Novelty**: is it genuinely new, given the verified findings? Already proved, refuted, or posed in the literature counts as not new.
- **Plausibility**: is it likely to be true?
- **Clarity**: is it precise enough to work on as stated?
- **Ambition**: how substantial would a resolution be?
- **Insight**: what would the community learn from settling it, whether it turns out true or false? Would working on it reveal new connections, mechanisms, or ways of thinking about the area?

For each consideration, say which conjecture is stronger: A, B, or tied.

# Output

Reply with one JSON object and nothing else:

```json
{
  "considerations": {"significance": "A | B | tied", "novelty": "A | B | tied", "plausibility": "A | B | tied",
                     "clarity": "A | B | tied", "ambition": "A | B | tied", "insight": "A | B | tied"},
  "same_conjecture": "yes | no",
  "choice": "A | B",
  "main_reason": "the main mathematical reason for your choice, in one to three sentences",
  "explanation": "a short paragraph comparing the two conjectures"
}
```
\end{promptbox}

\section{Seed Papers}
\label{app:seeds}

Table~\ref{tab:seeds} lists the 30 seed papers. Twelve were selected by a collaborating mathematician, an author of each, as the papers they felt most confident assessing; the remaining 18 were selected from Math-Graph for accessible full text, an original theorem, and statements understandable with graduate-level background in discrete mathematics and theoretical computer science.

\begin{table*}[t]
\centering
\small
\begin{tabular}{p{0.58\textwidth}llc}
\hline
Title & arXiv & Source & Experiments \\
\hline
Local-global principles for torsors over arithmetic curves & 1108.3323v5 & Expert & 1, 2 \\
Transcendental splitting fields of division algebras & 2203.15951v2 & Expert & 1, 2 \\
Index reduction for Brauer classes via stable sheaves (with an appendix by Bhargav Bhatt) & 0706.1072v3 & Expert & 1, 2 \\
Weierstrass preparation and algebraic invariants & 1109.6362v2 & Expert & 1, 2 \\
The Clifford algebra of a finite morphism & 1509.07195v4 & Expert & 1, 2 \\
On rational connectedness and parametrization of finite Galois extensions & 2502.15674v1 & Expert & 1 \\
Brauertsch fields & 2305.06464v1 & Expert & 1, 2, 3 \\
Local-global Galois theory of arithmetic function fields & 1710.03635v3 & Expert & 1 \\
Refinements to patching and applications to field invariants & 1404.4349v3 & Expert & 1, 2 \\
Local-global principles for Galois cohomology & 1208.6359v2 & Expert & 1, 2 \\
Patching subfields of division algebras & 0904.1594v2 & Expert & 1, 2 \\
Applications of patching to quadratic forms and central simple algebras & 0809.4481v2 & Expert & 1, 2 \\
How many random edges make an almost-Dirac graph Hamiltonian? & 2410.14447v1 & Math-Graph & 1, 2, 3 \\
Counting independent sets in regular graphs with bounded independence number & 2410.19959v1 & Math-Graph & 1, 2 \\
Inversion diameter and treewidth & 2407.15384v4 & Math-Graph & 1, 2 \\
Backbone coloring for graphs with degree 4 & 2409.10201v1 & Math-Graph & 1, 2, 3 \\
Extending Stable and Popular Matching Algorithms from Bipartite to Arbitrary Instances & 2409.16173v3 & Math-Graph & 1, 2, 3 \\
Computational complexity of the recoverable robust shortest path problem in acyclic digraphs under interval budgeted uncertainty & 2410.09425v3 & Math-Graph & 1, 2, 3 \\
Pseudo-Deterministic Construction of Irreducible Polynomials over Finite Fields & 2410.04071v1 & Math-Graph & 1, 2 \\
On Optimal Testing of Linearity & 2411.14431v1 & Math-Graph & 1, 2 \\
On maximal almost balanced non-overlapping codes and non-overlapping codes with restricted run-lengths & 2410.18458v1 & Math-Graph & 1, 2 \\
Colorful Intersections and Tverberg Partitions & 2403.14909v1 & Math-Graph & 1, 2, 3 \\
About almost covering subsets of the hypercube & 2409.10573v4 & Math-Graph & 1, 2 \\
Bounds for Greedy $B_h$-sets & 2312.10910v3 & Math-Graph & 1, 2 \\
On large Sidon sets & 2411.12911v3 & Math-Graph & 1, 2, 3 \\
Wythoff's Nim with Finite Alterations & 2408.02851v2 & Math-Graph & 1, 2, 3 \\
Counterexamples to two conjectures on mean color numbers of graphs & 2405.01890v2 & Math-Graph & 1, 2 \\
Optical orthogonal codes from a combinatorial perspective & 2411.06955v2 & Math-Graph & 1, 2 \\
Applications of Sparse Hypergraph Colorings & 2406.01499v1 & Math-Graph & 1, 2, 3 \\
Bounds on the Treewidth of Level-k Rooted Phylogenetic Networks & 2411.13380v1 & Math-Graph & 1, 2 \\
\hline
\end{tabular}
\caption{Seed papers. Source: Expert = selected by the collaborating mathematician; Math-Graph = selected from the Math-Graph dataset.}
\label{tab:seeds}
\end{table*}

\end{document}